# ToTMNet: FFT-Accelerated Toeplitz Temporal Mixing Network for Lightweight Remote Photoplethysmography

Vladimir Frants, Sos Agaian, *Life Fellow, IEEE*, and Karen Panetta, *Member, IEEE*

*Abstract*—Remote photoplethysmography (rPPG) estimates a blood volume pulse (BVP) waveform from facial videos captured by commodity cameras. Although recent deep models improve robustness compared to classical signal-processing approaches, many methods increase computational cost and parameter count, and attention-based temporal modeling introduces quadratic scaling with respect to the temporal length. This paper proposes ToTMNet, a lightweight rPPG architecture that replaces temporal attention with an FFT-accelerated Toeplitz temporal mixing layer. The Toeplitz operator provides full-sequence temporal receptive field using a linear number of parameters in the clip length and can be applied in near-linear time using circulant embedding and FFT-based convolution. ToTMNet integrates the global Toeplitz temporal operator into a compact gated temporal mixer that combines a local depthwise temporal convolution branch with gated global Toeplitz mixing, enabling efficient long-range temporal filtering while only having 63k parameters. Experiments on two datasets, UBFC-rPPG (real videos) and SCAMPS (synthetic videos), show that ToTMNet achieves strong heart-rate estimation accuracy with a compact design. On UBFC-rPPG intra-dataset evaluation, ToTMNet reaches 1.055 bpm MAE with Pearson correlation 0.996. In a synthetic-to-real setting (SCAMPS→UBFC-rPPG), ToTMNet reaches 1.582 bpm MAE with Pearson correlation 0.994. Ablation results confirm that the gating mechanism is important for effectively using global Toeplitz mixing, especially under domain shift. The main limitation of this preprint study is the use of only two datasets; nevertheless, the results indicate that Toeplitz-structured temporal mixing is a practical and efficient alternative to attention for rPPG.

## I. INTRODUCTION

REMOTE photoplethysmography (rPPG) estimates cardiovascular dynamics from ordinary video by exploiting subtle, periodic skin-color variations caused by blood volume changes. In practice, rPPG aims to recover a blood volume pulse (BVP) waveform and related measures such as heart rate (HR), using common RGB cameras available on laptops, smartphones, and telemedicine devices. The contactless nature of rPPG makes it attractive for comfort-oriented monitoring and scenarios where wearable sensors are inconvenient. Early studies demonstrated that pulse-related signals can be recovered from consumer cameras under ambient illumination, motivating a wide research community around camera-based vital sign estimation [1]–[3].

Despite this promise, rPPG remains challenging because the useful physiological component is weak and easily dominated by motion, illumination changes, compression artifacts, and variations in skin tone and camera response. Classical rPPG pipelines therefore combine careful signal processing with hand-crafted priors. Representative examples include chrominance-based methods that reduce sensitivity to motion and illumination changes [4], and more general analyses that formalize the optical and physiological factors behind robust pulse extraction and lead to principled algorithm design choices [5]. These approaches remain strong baselines in constrained settings, but they often degrade when conditions differ from the assumptions used during derivation.

To improve robustness and reduce manual engineering, recent work increasingly uses data-driven learning. Convolutional neural networks (CNNs) and spatiotemporal architectures learn to map video inputs to rPPG signals end-to-end or via intermediate representations. Early examples include attention-based dual-stream designs for physiological measurement [6], and spatiotemporal networks that directly model temporal dynamics in face videos [7]. For practical deployment, several architectures explicitly target efficiency, e.g., using temporal shift mechanisms and multi-task learning to reduce computational cost while maintaining accuracy [8]. In parallel, unsupervised and weakly supervised objectives have been explored to reduce dependence on synchronized ground-truth signals, which are often expensive to collect at scale [9], [10].

More recently, transformer-style models have been introduced to better capture long-range temporal dependencies and quasi-periodic structure. PhysFormer and its extensions demonstrate that temporal-difference tokenization and attention mechanisms can be effective for rPPG representation learning across multiple datasets [11], [12]. At the same time, multiple papers note that standard attention has quadratic cost in sequence length, which can force coarse tokenization and increase sensitivity to noise. This has motivated designs that explicitly exploit periodicity or restructure attention and feature aggregation, such as periodic sparse attention for rPPG [13] and more general multidimensional attention mechanisms that jointly consider spatial, temporal, and channel interactions [14]. These developments underline an ongoing trend: rPPG models benefit from stronger temporal modeling, but practical use still requires compute- and parameter-efficient designs.

In this work, we focus on the efficiency–accuracy trade-off for rPPG signal recovery. We propose ToTMNet, a Toeplitz Temporal Mixing Network that replaces heavyweight temporal modeling components with structured Toeplitz temporal mixing layers. A Toeplitz matrix is a structured matrix with constant diagonals; in temporal settings it naturally represents shift-invariant interactions and can be implemented efficiently. Inspired by recent results showing that Toeplitz-structured layers can serve as powerful and efficient token-mixing operators for sequence modeling [15], we adapt this idea to the rPPG setting and design a compact architecture that preserves competitive performance while significantly reducing parameter count. We validate ToTMNet on two datasets: UBFC-rPPG, a commonly used real-video benchmark [16], and SCAMPS, a high-fidelity synthetic dataset designed for camera-based physiological measurement with precise labels and controlled variability [17].

Contributions:
- We propose ToTMNet, a lightweight rPPG architecture based on Toeplitz temporal mixing layers, designed to reduce parameter count while maintaining competitive rPPG reconstruction and HR estimation performance.
- We describe an efficient implementation strategy for Toeplitz temporal mixing suitable for end-to-end training and long temporal windows.
- We evaluate ToTMNet on two public datasets (UBFC-rPPG and SCAMPS) and compare against representative strong baselines, highlighting the parameter–accuracy trade-off.

The rest of the paper is organised as follows. Section II reviews related work in rPPG and efficient temporal modeling. Section III presents the proposed ToTMNet architecture and the Toeplitz temporal mixing formulation. Section IV describes experimental protocol, datasets, metrics, and quantitative and qualitative results. Section V concludes the paper and discusses limitations and future work.

## II. Previous Work

Remote photoplethysmography (rPPG) estimates physiological waveforms from subtle, temporally coherent changes in skin appearance recorded by a camera. Early studies demonstrated that a camera can recover pulse information under ambient illumination, initially using simple channel selection (often the green channel) and later employing blind source separation to isolate the pulsatile component from motion and illumination variation [1], [2]. Building on these foundations, a large body of work proposed signal-processing pipelines that combine (i) region-of-interest (ROI) aggregation, (ii) color-space projections that amplify blood-volume related variations, and (iii) temporal filtering and peak analysis. Notable examples include chrominance-based formulations designed to improve robustness to illumination changes [4], motion-robust variants exploiting pulse-signature constraints [18], and standardized formulations of algorithmic design choices (e.g., the POS family) that clarified how color mixing and normalization influence stability in practical conditions [5].

Deep learning methods increasingly replaced hand-designed projections with learned spatiotemporal representations, motivated by the difficulty of preserving rPPG fidelity under non-stationary motion, heterogeneous lighting spectra, and camera processing artifacts. A widely adopted direction is to learn attention-like spatial weighting over the face and to regress a physiological waveform end-to-end from video, exemplified by convolutional attention networks such as DeepPhys [6]. Spatiotemporal backbones further improved temporal modeling capacity by using 3D convolutions or recurrent-style temporal aggregation; PhysNet is a representative early end-to-end spatiotemporal approach that explicitly targets waveform reconstruction rather than only heart-rate regression [7]. Efficiency and deployability have also been emphasized, for example via temporal-shift based designs that provide strong temporal modeling at low cost (e.g., MTTS-CAN) [8] and architectures that remove extensive preprocessing while maintaining competitive accuracy (e.g., EfficientPhys) [19]. These methods collectively show that learned temporal modeling is beneficial, but they also highlight practical trade-offs between model size, temporal receptive field, and robustness across domains.

More recently, transformer-style architectures were explored to better capture long-range spatiotemporal interactions in facial videos. PhysFormer introduces temporal-difference guided attention and frequency-domain supervision to emphasize quasi-periodic rPPG components and mitigate overfitting to nuisance variation [11]. Subsequent designs investigated alternative attention factorization and dimension-wise coupling, such as matrix-factorization approaches that aim to reduce the cost of multidimensional attention while preserving expressivity [14]. In parallel, work explicitly encoding periodic structure into attention has been proposed; RhythmFormer, for example, uses periodic sparse attention intended to exploit the periodicity of physiological signals while reducing irrelevant attention computation [13]. Collectively, these methods emphasize that rPPG contains long-range structure (e.g., periodicity with mild non-stationarity) that can be exploited, but they also underscore that quadratic-cost attention can be mismatched to long temporal windows commonly required for stable pulse estimation.

Alongside supervised training, there is growing interest in reducing reliance on synchronized contact sensors, since acquiring clean ground-truth photoplethysmogram (PPG) signals is costly and can limit dataset scale. Self-supervised and unsupervised methods use weak priors (e.g., temporal smoothness and plausible frequency bands) and contrastive objectives to learn representations from unlabeled video. Gideon and Stent proposed a contrastive-learning formulation for rPPG from unlabelled video, demonstrating that meaningful physiological representations can be learned without direct waveform supervision [20]. Contrast-Phys further develops this direction via spatiotemporal contrast and multiple rPPG predictions per video, yielding improved unsupervised performance and competitive efficiency [9]. This line of work suggests that rPPG estimation benefits from inductive biases that enforce physiological regularities even when explicit labels are unavailable.

Because reported results can be sensitive to preprocessing choices, evaluation protocols, and dataset splits, reproducible benchmarking has become an important topic. rPPG-Toolbox provides an end-to-end framework that includes dataset handling, implementations of multiple neural and unsupervised baselines, and systematic evaluation pipelines, helping standardize comparisons and reduce hidden implementation variability across studies [21]. Such toolchains also make it easier to assess the effect of architectural changes in isolation, which is particularly relevant when claims involve parameter efficiency or structured parameterizations.

A separate but closely related body of research focuses on reducing parameters and/or accelerating temporal modeling through structured computation. In sequence modeling, the transformer self-attention operator is powerful but expensive, motivating many subquadratic alternatives. These include kernel-based linear attention approximations (e.g., Performer) [22], low-rank projections of attention (e.g., Linformer) [23], sparse or block-sparse attention patterns for long documents (e.g., Longformer) [24], locality-sensitive hashing based attention (e.g., Reformer) [25], and hardware-aware kernels for exact attention that improve memory and throughput (e.g., FlashAttention) [26]. In parallel, structured state-space models (SSMs) provide another route to long-context modeling with favorable scaling: S4 uses a structured parameterization of linear dynamical systems to model long-range dependencies efficiently [27], while later variants and related designs (e.g., S5, Mamba) further simplify parameterization or introduce input-dependent state updates for improved performance and scaling [28], [29]. Long-convolution operators, such as Hyena, represent another alternative that targets efficient long-context mixing without explicit attention [30], and Fourier-based token mixing (e.g., FNet) replaces attention with fast linear transforms to reduce cost while retaining reasonable quality for certain tasks [31]. These approaches share a common theme: they inject architectural inductive bias and exploit algebraic structure to replace generic dense mixing with parameter-efficient, computationally efficient operators.

Structured matrices and structured linear layers are a particularly relevant lens for parameter reduction, because they directly constrain weight matrices to admit compact representations and fast multiplication. Circulant projections replace dense fully-connected layers with circulant structure, enabling fast multiplication via the fast Fourier transform (FFT) and reducing storage from quadratic to linear in the layer width [32]. Block-circulant weight matrices extend this idea to broader network components and hardware-friendly acceleration strategies (e.g., CirCNN) [33]. More generally, the low displacement rank (LDR) framework unifies multiple families of structured matrices (including Toeplitz-like and circulant-like constructions) and provides a systematic way to learn compact transforms with fast algorithms; structured transforms based on LDR have been used to build small-footprint models [34] and to learn more expressive compressed transforms while retaining quasi-linear multiplication complexity [35]. Other structured parameterizations target fast algorithms for common linear transforms and compressions (e.g., butterfly factorizations) [36], and randomized structured transforms (e.g., Fastfood) were early demonstrations that structured matrices can provide strong practical gains in memory and speed while approximating dense mappings [37]. Within long-sequence modeling, Toeplitz structure has been used directly as an efficient relative-position mixing mechanism (Toeplitz Neural Network), showing that Toeplitz matrix–vector products can provide log-linear complexity while preserving competitive sequence modeling performance [15]. Related theoretical work links structured masking and positional mechanisms in transformers to block-Toeplitz patterns and graph-based views, providing a mathematical foundation for scalable masked attention with structural inductive bias [38]. These results collectively motivate structured temporal mixing as a principled path to compact models: Toeplitz and Toeplitz-like operators encode shift-structured interactions, which can align well with signals that exhibit regular temporal dynamics.

Finally, alternative parameter-sharing strategies have been explored for multi-channel signals through algebras beyond the real numbers. Parameterized hypercomplex convolutions (PHNNs) demonstrate that structured channel mixing rules can reduce parameters while preserving performance across modalities, and they provide a flexible framework that can be adapted to different dimensionalities and domains [39]. In the broader signal-processing and learning literature, complex-valued neural networks have been surveyed extensively, with emphasis on how structured representations can better match naturally oscillatory or phase-sensitive data [40]. While such approaches are not yet standard in mainstream rPPG pipelines, they reinforce the general principle that constraining interactions across channels and time through mathematically structured operators can improve efficiency and sometimes generalization by embedding useful inductive biases rather than relying solely on over-parameterized dense layers.

## III. METHOD

This section presents ToTMNet. The design consists of (i) a framewise spatial stem that converts each frame into a compact feature vector, (ii) a stack of gated local–global temporal mixer blocks that combine short-range temporal convolution with global Toeplitz mixing, and (iii) a per-frame regression head that outputs the rPPG waveform.

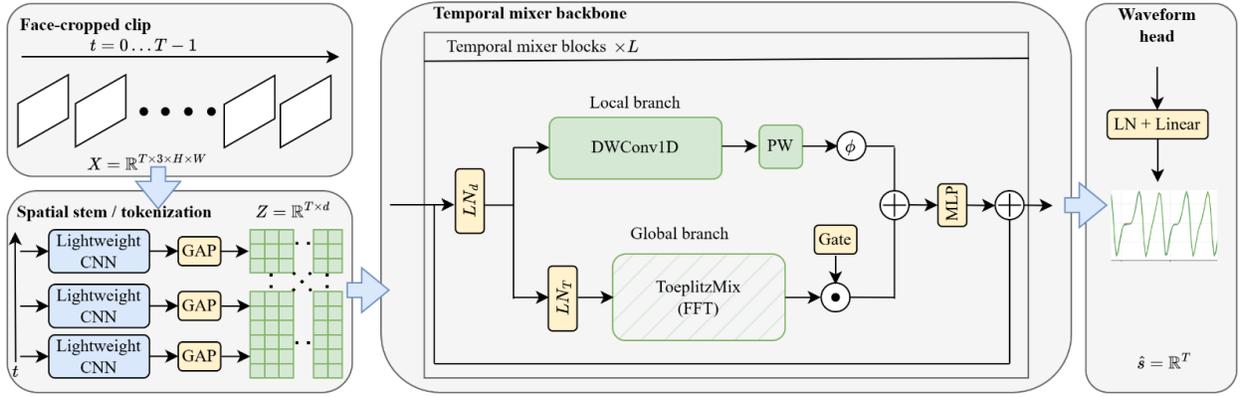

**Fig. 1.** Overview of ToTMNet. A face-cropped RGB clip $\mathbf{X} \in \mathbb{R}^{B \times T \times 3 \times H \times W}$ is tokenized framewise by a lightweight spatial stem (CNN + global average pooling) to produce temporal embeddings $\mathbf{Z} \in \mathbb{R}^{B \times T \times d}$. A stack of $L$ temporal mixer blocks models dynamics by combining a local depthwise 1D temporal convolution branch (DWConv1D + pointwise projection) with a global Toeplitz temporal mixing branch computed via FFT; a sigmoid gate modulates the global contribution before residual fusion and a lightweight MLP. A layer-normalized linear head regresses the per-frame rPPG waveform $\hat{\mathbf{s}} \in \mathbb{R}^{B \times T}$. *GAP*: global average pooling; *LN*: layer normalization; *PW*: pointwise projection.

*A. Problem formulation and notation*

Let a batch of face-cropped video clips be $\mathbf{X} \in \mathbb{R}^{B \times T \times C \times H \times W}$, where $B$ is the batch size, $T$ is the number of frames per clip, $C$ is the number of channels (in this work, $C = 3$ for RGB-based inputs), and $H \times W$ is the spatial resolution. We index time by $t \in \{0, \ldots, T-1\}$. The goal is to predict a per-frame rPPG waveform $\hat{\mathbf{s}} \in \mathbb{R}^{B \times T}$, that matches a reference blood volume pulse (BVP) waveform in both temporal shape and dominant frequency content.

Below, $d$ denotes an embedding dimension and $L$ a stack depth. For intermediate sequences, we use the convention that tensors in $\mathbb{R}^{B \times T \times d}$ are arranged with time as the second dimension. The elementwise product is denoted by $\odot$, and $\sigma(\cdot)$ denotes the logistic sigmoid. We denote layer normalization across the feature dimension by $\mathrm{LN}_d(\cdot)$, applied independently at each time step:

$$\bigl(\mathrm{LN}_d(\mathbf{H})\bigr)_{b,t} = \mathrm{LN}(\mathbf{H}_{b,t}) \quad \text{for } \mathbf{H} \in \mathbb{R}^{B \times T \times d} \tag{1}$$

We also use a temporal layer normalization $\mathrm{LN}_T(\cdot)$, applied independently to each feature channel trace across time:

$$\bigl(\mathrm{LN}_T(\mathbf{H})\bigr)_{b,j} = \mathrm{LN}(\mathbf{H}_{b,j}) \quad \text{for } j \in \{0, \ldots, d-1\} \tag{2}$$

The ToTMNet workflow is depicted in Fig. 1: the clip $\mathbf{X}$ is mapped to temporal tokens $\mathbf{Z} \in \mathbb{R}^{B \times T \times d}$ by the framewise spatial stem $f_{\mathrm{sp}}$ (applied independently to each frame), then $L$ temporal mixer blocks $f_{\mathrm{blk}}$ produce $\mathbf{H}^L$, serving as an input of the regression head $f_{\mathrm{head}}$ to estimate $\hat{\mathbf{s}}$:

$$\mathbf{Z} = f_{\mathrm{sp}}(\mathbf{X}) \tag{3}$$
$$\mathbf{H}^{(0)} = \mathbf{Z} \tag{4}$$
$$\mathbf{H}^{(\ell+1)} = f_{\mathrm{blk}}\bigl(\mathbf{H}^{(\ell)}\bigr), \ell = 0, \ldots, L-1 \tag{5}$$
$$\hat{\mathbf{s}} = f_{\mathrm{head}}\bigl(\mathbf{H}^{(L)}\bigr) \tag{6}$$

The spatial stem $f_{\mathrm{sp}}$ is applied independently to each frame. Let $\mathbf{x}_{b,t} \in \mathbb{R}^{C \times H \times W}$ be the $t$-th frame of clip $b$. The stem produces an embedding $\mathbf{z}_{b,t} \in \mathbb{R}^d$, then stacked across time:

$$\mathbf{Z} = \bigl[\mathbf{z}_{b,0}, \mathbf{z}_{b,1}, \ldots, \mathbf{z}_{b,T-1}\bigr] \in \mathbb{R}^{B \times T \times d} \tag{7}$$

The stem is designed to be lightweight (e.g., based on small convolutions and global average pooling), reducing spatial dimensions early so that subsequent computation focuses on temporal modeling, consistent with the weak but structured temporal nature of rPPG.

The model predicts one scalar per frame from the final temporal features $\mathbf{H}^{(L)} \in \mathbb{R}^{B \times T \times d}$. A normalized linear head produces

$$\hat{s}_{b,t} = \mathbf{w}_{\mathrm{out}}^\top \mathrm{LN}_d\bigl(\mathbf{H}^{(L)}\bigr)_{b,t} + b_{\mathrm{out}} \tag{8}$$

where $\mathbf{w}_{\mathrm{out}} \in \mathbb{R}^d$ and $b_{\mathrm{out}} \in \mathbb{R}$. Collecting all $\hat{s}_{b,t}$ yields $\hat{\mathbf{s}} \in \mathbb{R}^{B \times T}$.

The overall ToTMNet inference pipeline is summarized in Algorithm 1, and the FFT-based ToeplitzMix routine used to implement the global Toeplitz temporal mixing operator is detailed in Algorithm 2.

The Toeplitz operator imposes a shift-structured temporal inductive bias ($A_{m,n}$ depends only on $m - n$), which is consistent with global filtering commonly used in physiological signal processing. Compared to attention-based temporal mixing with $O(T^2)$ complexity, Toeplitz mixing admits $O(T \log T)$ evaluation via FFT and requires only $2T - 1$ parameters (excluding optional bias), making it attractive for lightweight rPPG modeling over longer clips.

*B. Gated local-global temporal mixer block*

Each temporal block combines a local temporal convolution branch (capturing short-range dynamics) and a global Toeplitz temporal mixing branch (capturing long-range dependencies). Let $\mathbf{H} \in \mathbb{R}^{B \times T \times d}$ be the block input (dropping the layer index for clarity). The block proceeds as follows:

1) **Normalization**

A feature-wise normalization is first applied: $\tilde{\mathbf{H}} = \mathrm{LN}_d(\mathbf{H})$.

2) **Local temporal branch**

The local branch applies a depthwise temporal convolution of kernel size $K$ (in our setting, $K = 5$), followed by a pointwise (channel-mixing) projection and a nonlinearity $\phi(\cdot)$ (SiLU):

$$\mathbf{U} = \mathrm{PW}\Bigl(\phi\bigl(\mathrm{DWConv}_K(\tilde{\mathbf{H}})\bigr)\Bigr), \quad \mathbf{U} \in \mathbb{R}^{B \times T \times d} \tag{9}$$

**ALGORITHM 1. FORWARD PASS**

**Require:** input clip $\mathbf{X}$; number of blocks $L$; spatial stem $f_{\text{sp}}$; head $f_{\text{head}}$; per-block parameters $\{\mathbf{c}^{(\ell)}, \mathbf{r}^{(\ell)}, \mathbf{W}_g^{(\ell)}, \mathbf{b}_g^{(\ell)}, \text{MLP}^{(\ell)}\}_{\ell=0}^{L-1}$
**Ensure:** predicted waveform $\hat{\mathbf{s}}$
1:     $\mathbf{Z} \leftarrow f_{\text{sp}}(\mathbf{X})$
2:     $\mathbf{H} \leftarrow \mathbf{Z}$
3:     **for** $\ell = 0$ **to** $L - 1$ **do**
4:        $\tilde{\mathbf{H}} \leftarrow \text{LN}_d(\mathbf{H})$
5:        $\mathbf{U} \leftarrow \text{Local}^{(\ell)}(\tilde{\mathbf{H}})$
6:        $\mathbf{V} \leftarrow \text{Toeplitz}^{(\ell)}(\tilde{\mathbf{H}})$
7:        $\mathbf{G} \leftarrow \sigma\left(\tilde{\mathbf{H}}(\mathbf{W}_g^{(\ell)})^\top + \mathbf{b}_g^{(\ell)}\right)$
8:        $\bar{\mathbf{H}} \leftarrow \mathbf{H} + \mathbf{U} + \mathbf{G} \odot \mathbf{V}$
9:        $\mathbf{H} \leftarrow \bar{\mathbf{H}} + \text{MLP}^{(\ell)}(\text{LN}_d(\bar{\mathbf{H}}))$
10:    **end for**
11:    $\hat{\mathbf{s}} \leftarrow f_{\text{head}}(\mathbf{H}) = \text{Linear}_{\text{out}}(\text{LN}_d(\mathbf{H}))$
12:    **return** $\hat{\mathbf{s}}$

3) **Global Toeplitz branch**

The global branch applies Toeplitz temporal mixing along the time dimension. Prior to mixing, a temporal normalization is applied per channel:

$$\mathbf{V} = \text{ToTM}\left(\text{LN}_T(\tilde{\mathbf{H}})\right), \mathbf{V} \in \mathbb{R}^{B \times T \times d} \quad (10)$$

4) **Sigmoid gate and residual fusion**

A learned gate modulates the global contribution:

$$\mathbf{G} = \sigma(\tilde{\mathbf{H}}\mathbf{W}_g^\top + \mathbf{b}_g), \qquad \mathbf{W}_g \in \mathbb{R}^{d \times d}, \mathbf{b}_g \in \mathbb{R}^d \quad (11)$$

where the affine map is applied independently at each time step. The fused residual update is

$$\bar{\mathbf{H}} = \mathbf{H} + \mathbf{U} + \mathbf{G} \odot \mathbf{V} \quad (12)$$

This gating allows the model to adaptively balance local (convolutional) and global (Toeplitz) temporal information.

5) **Position-wise feed-forward network**

A lightweight per-time-step MLP refines features with a second residual connection. Let the MLP hidden dimension be $h$. Then

$$\text{MLP}(\mathbf{x}) = \mathbf{W}_2 \phi(\mathbf{W}_1 \mathbf{x}), \mathbf{W}_1 \in \mathbb{R}^{h \times d}, \mathbf{W}_2 \in \mathbb{R}^{d \times h} \quad (13)$$

and the block output is:

$$\mathbf{H}_{\text{out}} = \bar{\mathbf{H}} + \text{MLP}(\text{LN}_d(\bar{\mathbf{H}})) \quad (14)$$

*C. Toeplitz temporal mixing operator*

This subsection defines the Toeplitz temporal operator used in the global branch and its efficient evaluation.

1) **Toeplitz structure and parameterization**

A matrix $\mathbf{A} \in \mathbb{R}^{T \times T}$ is Toeplitz if its entries are constant along diagonals:

$$A_{m,n} = \tau_{m-n}, \qquad m, n \in \{0, \dots, T-1\} \quad (15)$$

for a set of lag parameters $\{\tau_{-(T-1)}, \dots, \tau_{T-1}\}$. Hence, a $T \times T$ Toeplitz matrix has $2T - 1$ degrees of freedom (excluding any optional bias).

Equivalently, $\mathbf{A}$ is determined by its first column $\mathbf{c} \in \mathbb{R}^T$ and first row $\mathbf{r} \in \mathbb{R}^T$, with the constraint $r_0 = c_0$:

$$\mathbf{c} = [A_{0,0}, A_{1,0}, \dots, A_{T-1,0}]^\top \quad (16)$$

$$\mathbf{r} = [A_{0,0}, A_{0,1}, \dots, A_{0,T-1}] \quad (17)$$

**ALGORITHM 2. FFT-BASED TOEPLITZMIX**

**Require:** $\mathbf{Q} \in \mathbb{R}^{B \times T \times d}$; $\mathbf{c}, \mathbf{r} \in \mathbb{R}^T$.
**Ensure:** $\mathbf{V} \in \mathbb{R}^{B \times T \times d}$
1:     $r_0 \leftarrow c_0$; $L \leftarrow 2T - 1$
2:     $\boldsymbol{\kappa} \leftarrow [\mathbf{c} \; ; \; \text{rev}(\mathbf{r}_{1:T-1})] \in \mathbb{R}^L$
3:     **for all** $b \in \{1, \dots, B\}, j \in \{0, \dots, d-1\}$ **do**
4:        $\mathbf{x} \leftarrow \mathbf{Q}_{b,:,j} \in \mathbb{R}^T$
5:        $\mathbf{x}_{\text{pad}} \leftarrow [\mathbf{x} \; ; \; \mathbf{0}_{T-1}] \in \mathbb{R}^L$
6:        $\mathbf{z} \leftarrow \text{IFFT}\left(\text{FFT}(\boldsymbol{\kappa}) \odot \text{FFT}(\mathbf{x}_{\text{pad}})\right) \in \mathbb{R}^L$
7:        $\mathbf{V}_{b,:,j} \leftarrow \mathbf{z}_{0:T-1} \in \mathbb{R}^T$
8:     **end for**
9:     **return** $\mathbf{V}$

The Toeplitz matrix is then

$$\mathbf{A}(\mathbf{c}, \mathbf{r}) = \begin{bmatrix} c_0 & r_1 & r_2 & \cdots & r_{T-1} \\ c_1 & c_0 & r_1 & \cdots & r_{T-2} \\ c_2 & c_1 & c_0 & \cdots & r_{T-3} \\ \vdots & \vdots & \vdots & \ddots & \vdots \\ c_{T-1} & c_{T-2} & c_{T-3} & \cdots & c_0 \end{bmatrix} \quad (18)$$

2) **Application to temporal tokens**

Given a token sequence $\mathbf{Q} \in \mathbb{R}^{B \times T \times d}$, Toeplitz mixing applies the same temporal operator to each feature channel. For each batch element $b$, let $\mathbf{Q}_b \in \mathbb{R}^{T \times d}$. The mixed output $\mathbf{V}_b \in \mathbb{R}^{T \times d}$ is

$$\mathbf{V}_b = \mathbf{A}(\mathbf{c}, \mathbf{r}) \mathbf{Q}_b \quad (19)$$

This shared-operator design makes the number of Toeplitz parameters independent of $d$, which is advantageous for compact models.

3) **FFT-based evaluation via circulant embedding**

Direct multiplication by $\mathbf{A}$ scales as $O(T^2)$. Toeplitz multiplication can be evaluated in $O(T \log T)$ using an FFT-based circulant embedding of size $L = 2T - 1$.

Let's define the length-$L$ vector

$$\boldsymbol{\kappa} = [c_0, c_1, \dots, c_{T-1}, r_{T-1}, r_{T-2}, \dots, r_1] \in \mathbb{R}^L,$$

and for $\mathbf{x} \in \mathbb{R}^T$ define the zero-padded vector $\mathbf{x}_{\text{pad}} \in \mathbb{R}^L$ by appending $T - 1$ zeros. Let $\mathcal{F}_L$ and $\mathcal{F}_L^{-1}$ denote the length-$L$ discrete Fourier transform and its inverse, respectively. Then

$$\mathbf{z} = \mathcal{F}_L^{-1}\left(\mathcal{F}_L(\boldsymbol{\kappa}) \odot \mathcal{F}_L(\mathbf{x}_{\text{pad}})\right) \in \mathbb{R}^L \quad (20)$$

and the Toeplitz product is obtained as the first $T$ entries:

$$\mathbf{y} = \mathbf{z}_{0:T-1} = \mathbf{A}(\mathbf{c}, \mathbf{r}) \mathbf{x} \quad (21)$$

Applying this procedure columnwise to $\mathbf{Q}_b \in \mathbb{R}^{T \times d}$ yields $\mathbf{V}_b$. The resulting per-block cost for global mixing scales as $O(T \log T)$ per channel (and can be batched across channels), providing global temporal receptive field without attention's quadratic scaling.

*D. Loss function*

To encourage agreement in both time and frequency domains, ToTMNet can be trained with a weighted combination of losses. Let $\mathbf{s} \in \mathbb{R}^{B \times T}$ denote the reference waveform and $\hat{\mathbf{s}} \in \mathbb{R}^{B \times T}$ the prediction. The objective is

$$\mathcal{L} = \lambda_{\text{mse}} \mathcal{L}_{\text{mse}} + \lambda_\rho \mathcal{L}_\rho + \lambda_{\text{spec}} \mathcal{L}_{\text{spec}} \quad (22)$$

where

$$\lambda_{\text{mse}}, \lambda_\rho, \lambda_{\text{spec}} \geq 0.$$

Time-domain error:

$$\mathcal{L}_{\text{mse}} = \frac{1}{BT} \sum_{b=1}^{B} \sum_{t=0}^{T-1} (\hat{s}_{b,t} - s_{b,t})^2 \quad (23)$$

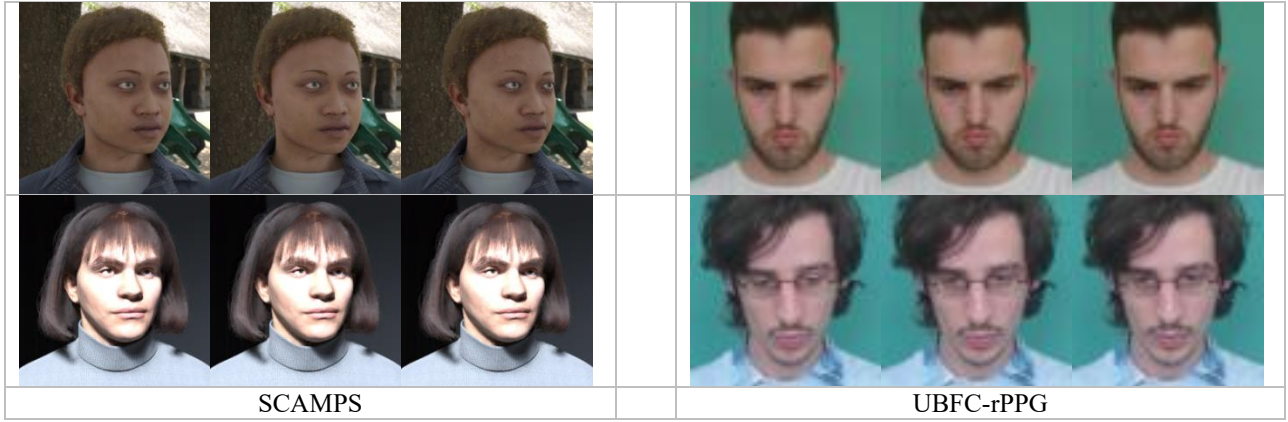

| SCAMPS | UBFC-rPPG |

**Fig. 2.** Example face regions of interest used as network input. Representative cropped face patches from SCAMPS (synthetic; left) and UBFC-rPPG (real; right) illustrating typical appearance, illumination, and rendering differences across domains. Each row shows consecutive frames from a subject.

Correlation agreement: for each $b$, let mean-centered signals be:

$$\hat{s}'_{b,t} = \hat{s}_{b,t} - \frac{1}{T}\sum_{u=0}^{T-1}\hat{s}_{b,u} \quad (24)$$

$$s'_{b,t} = s_{b,t} - \frac{1}{T}\sum_{u=0}^{T-1} s_{b,u} \quad (25)$$

The Pearson correlation coefficient is

$$\rho_b = \frac{\sum_{t=0}^{T-1}\hat{s}'_{b,t}s'_{b,t}}{\sqrt{\left(\sum_{t=0}^{T-1}(\hat{s}'_{b,t})^2\right)\left(\sum_{t=0}^{T-1}(s'_{b,t})^2\right)+\epsilon}} \quad (26)$$

and the loss term is $\mathcal{L}_\rho = \frac{1}{B}\sum_{b=1}^{B}(1-\rho_b)$, with small $\epsilon > 0$.

Spectral magnitude agreement. Let $|\text{STFT}(\cdot)|$ denote the magnitude of a short-time Fourier transform, optionally restricted to a physiologically plausible frequency band. A spectral loss can be written as:

$$\mathcal{L}_{\text{spec}} = \frac{1}{|\Omega|}\sum_{(b,\omega,\tau)\in\Omega}\left\|\left|\text{STFT}(\hat{\mathbf{s}})_{b,\omega,\tau}\right| - \left|\text{STFT}(\mathbf{s})_{b,\omega,\tau}\right|\right\|_p$$

where $\Omega$ indexes selected time–frequency bins and $p \in \{1,2\}$.

## IV. Experiments

*A. Experimental setup*

We evaluate on two public datasets.

UBFC-rPPG contains real facial videos with synchronized reference pulse signals. We use the subject split with 33 subjects for training, 4 for validation, and 5 for testing. The median frame rate is 29.70 FPS and the median video duration is 68.18 s.

SCAMPS contains synthetic facial videos with ground-truth physiological signals. We use 2240 clips for training, 280 for validation, and 280 for testing. The frame rate is 30 FPS and the clip duration is 20 s. The reference BVP label is d_ppg in the dataset files.

Figure 2 shows example face crops / regions of interest from UBFC-rPPG and SCAMPS used for network input.

To isolate modeling effects, we evaluate ToTMNet and baselines under a matched preprocessing and evaluation pipeline based on rPPG-Toolbox conventions. Face crops are generated using the HC backend with a large face box coefficient of 1.5, with dynamic face detection disabled. Crops are resized to $72 \times 72$ pixels. Video clips are segmented into chunks of length $T = 180$ frames for training and evaluation, except for PhysFormer which is run with chunk length 160 in our configuration.

The preprocessing configuration DNSTD concatenates two representations, DiffNormalized and Standardized. When DNSTD is used, ToTMNet selects one half as the network input (three channels), while the waveform labels follow DiffNormalized, consistent with the toolbox configuration. Heart rate is computed from the predicted waveform using FFT-based spectral peak selection at sampling rate 30 Hz with full-window evaluation.

We compare against representative deep rPPG baselines provided by the toolbox: EfficientPhys, PhysNet, TS-CAN, and PhysFormer. For intra-dataset evaluation, we also report classical baselines including POS, CHROM, GREEN, and ICA.

ToTMNet uses a pyramid spatial stem and an embedding dimension $d = 32$. The temporal backbone uses three temporal mixer blocks. The local branch uses depthwise temporal convolution with kernel size 5. The MLP expansion ratio is 3.0 and dropout is 0.1. Toeplitz mixing uses the FFT implementation and a full window (no truncation) in the best-performing UBFC-rPPG and SCAMPS→UBFC-rPPG configuration.

We report heart-rate metrics and waveform metrics commonly used in rPPG evaluation. Heart-rate accuracy is summarized by mean absolute error (MAE, bpm), root mean squared error (RMSE, bpm), mean absolute percentage error (MAPE, %), and Pearson correlation ($\rho$) between predicted and reference heart rate. Waveform fidelity is reported by signal-to-noise ratio (SNR, dB) between predicted and reference waveforms.

Figure 3 shows example extracted rPPG waveforms (predicted vs ground truth) for UBFC-rPPG.

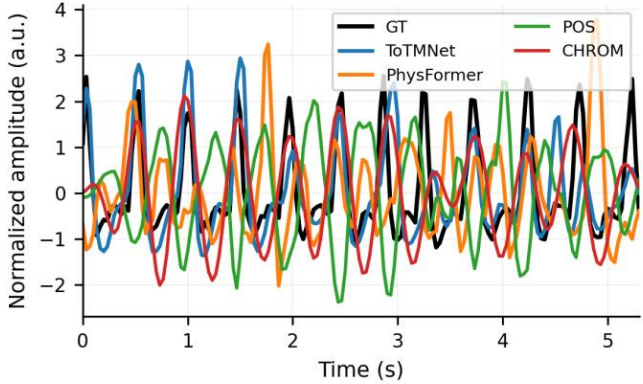

**Fig. 3.** Example extracted rPPG waveforms (time domain). Overlay of normalized waveform amplitude (a.u.) versus time for the ground-truth BVP (GT) and predicted rPPG signals from ToTMNet and representative baselines (e.g., PhysFormer, POS, CHROM) on a representative test window. Curves are shown after normalization for visual comparison of temporal agreement.

### B. Intra-dataset evaluation

Table I reports intra-dataset results on UBFC-rPPG. ToTMNet achieves the lowest MAE and RMSE among the compared methods in this evaluation setting. Compared to the deep baselines (EfficientPhys, PhysNet, TS-CAN), ToTMNet reduces MAE from 2.285 bpm to 1.055 bpm.

TABLE I
UBFC-rPPG INTRA-DATASET RESULTS

| Method | # parameters | MAE↓ | RMSE↓ | MAPE↓ | Pearson↑ | SNR↑ |
|---|---|---|---|---|---|---|
| GREEN | - | 30.694 | 42.803 | 42.803 | 0.199 | -9.916 |
| ICA | - | 21.905 | 34.141 | 34.141 | 0.356 | -8.011 |
| CHROM | - | 5.814 | 16.690 | 16.690 | 0.751 | -3.424 |
| POS | - | 4.733 | 15.692 | 15.692 | 0.778 | -2.602 |
| EfficientPhys | 2.163M | 3.431 | 4.467 | 4.181 | 0.994 | -1.771 |
| PhysNet | 0.769M | 5.326 | 7.802 | 7.030 | 0.986 | **-1.384** |
| TS-CAN | 2.229M | 2.285 | 3.871 | 3.871 | 0.992 | -2.078 |
| PhysFormer | 7.381M | 8.086 | 10.793 | 10.793 | 0.946 | -3.199 |
| ToTMNet | 0.063M | **1.055** | **2.358** | **2.358** | **0.996** | -1.387 |

An important observation is that several deep baselines produce identical HR metrics (MAE, RMSE, MAPE, Pearson) under FFT-based HR evaluation in this setup. This can occur when different waveform predictors yield the same dominant spectral peak, even if their waveforms differ in shape or noise level. The SNR values differ across these models, which suggests that waveform-level differences exist even when HR metrics coincide. For a journal version, it is useful to complement full-window FFT HR evaluation with additional waveform analyses or windowed evaluation protocols, so that differences in waveform quality are more visible.

Table II reports intra-dataset results on SCAMPS. ToTMNet achieves the lowest MAE and RMSE and the highest Pearson correlation among the listed deep baselines. PhysNet achieves a slightly lower MAPE and a marginally higher SNR, indicating that different models may trade off small differences in frequency accuracy and waveform fidelity on synthetic data.

TABLE II
SCAMPS INTRA-DATASET RESULTS

| Method | MAE↓ | RMSE↓ | MAPE↓ | Pearson↑ | SNR↑ |
|---|---|---|---|---|---|
| EfficientPhys | 1.387 | 7.251 | 2.789 | 0.971 | 2.203 |
| PhysNet | 0.979 | 6.651 | **1.611** | 0.976 | 4.448 |
| TS-CAN | 2.122 | 10.125 | 3.867 | 0.943 | 2.308 |
| PhysFormer | 1.821 | 9.575 | 3.695 | 0.949 | **4.458** |
| ToTMNet | **0.866** | **6.204** | 1.667 | **0.979** | 4.281 |

### C. Cross-dataset evaluation

Table III summarizes cross-dataset results between SCAMPS and UBFC-rPPG. The synthetic-to-real protocol (SCAMPS→UBFC-rPPG) is especially relevant when synthetic data is considered as a scalable training source. In this setting, ToTMNet achieves 1.582 bpm MAE and Pearson 0.994, improving over the deep baselines. Notably, EfficientPhys fails to transfer in this configuration, with a large error, while PhysNet and PhysFormer also degrade substantially compared to their intra-dataset performance.

The reverse direction (UBFC-rPPG→SCAMPS) is challenging for all methods, with high MAE values. ToTMNet achieves the lowest MAE in this direction, but the overall performance indicates a large domain gap. This asymmetry suggests that training on SCAMPS provides signals and variability that can transfer to UBFC-rPPG under this protocol, while training on UBFC-rPPG does not capture enough of the synthetic domain variability or the label characteristics needed to perform well on SCAMPS. This observation motivates future work on domain generalization and on using more diverse real datasets.

TABLE III
INTER-DATASET RESULTS

| Train → Test | Method | MAE↓ | RMSE↓ | MAPE↓ | Pearson↑ | SNR↑ |
|---|---|---|---|---|---|---|
| SCAMPS → UBFC-rPPG | EfficientPhys | 35.332 | 50.543 | 28.827 | -0.563 | -5.671 |
|  | PhysNet | 4.043 | 6.051 | 4.619 | 0.985 | -5.209 |
|  | TS-CAN | 2.285 | 3.871 | 3.071 | 0.992 | -3.907 |
|  | PhysFormer | 6.680 | 11.935 | 9.124 | 0.938 | **-3.580** |
|  | ToTMNet | **1.582** | **2.862** | **1.660** | **0.994** | -5.385 |
| UBFC-rPPG → SCAMPS | EfficientPhys | 25.777 | 41.101 | 24.394 | **0.421** | -5.852 |
|  | PhysNet | 24.459 | 35.691 | 27.677 | 0.143 | -6.314 |
|  | TS-CAN | 30.253 | 44.630 | 28.147 | 0.327 | -6.442 |
|  | PhysFormer | 23.956 | 32.839 | 27.271 | 0.179 | -7.210 |
|  | ToTMNet | **21.175** | **31.291** | **24.465** | 0.317 | **-5.753** |

Overall, these results suggest that Toeplitz-based global temporal mixing can provide a strong inductive bias for rPPG. The gains are most visible when global temporal context must be exploited efficiently, and when robustness to domain shift is required. At the same time, the limited dataset coverage in this preprint prevents strong general claims about in-the-wild robustness. Extending evaluation to additional real datasets is necessary for a journal-quality study.

### D. Ablation Study

The ablation study isolates the effect of the global Toeplitz temporal mixing and the gating mechanism. We compare three variants: a local-only model that uses only the depthwise temporal convolution branch; a model that adds Toeplitz mixing but injects it without gating; and the full ToTMNet model with gated Toeplitz mixing.

Table IV reports the ablation results on UBFC-rPPG intra-dataset evaluation. The local-only variant reaches 2.285 bpm MAE, which is comparable to several deep baselines under this evaluation setting. Adding Toeplitz mixing without gating does not improve performance; it slightly degrades MAE to 2.637 bpm. In contrast, the gated design improves MAE substantially to 1.055 bpm. This pattern indicates that global temporal mixing is not uniformly beneficial when applied in an unconditional manner. The rPPG signal is approximately periodic but the nuisance factors are non-stationary; a global operator can therefore propagate nuisance patterns across the clip if it is not controlled. The gate allows the model to suppress the global branch when the input segment contains unreliable temporal patterns and to emphasize it when the periodic component is strong.

TABLE IV
CORE ABLATION ON UBFC-rPPG INTRA-DATASET

| Variant | MAE↓ | RMSE↓ | MAPE↓ | Pearson↑ | SNR↑ |
|---|---|---|---|---|---|
| Local-only | 2.285 | 3.871 | 3.071 | 0.992 | -2.190 |
| No gate | 2.637 | 4.252 | 3.467 | 0.989 | -1.948 |
| ToTMNet | 1.055 | 2.358 | 1.188 | 0.996 | -1.387 |

Table V reports the same ablation under SCAMPS→UBFC-rPPG cross-dataset evaluation. This setting is more sensitive to mismatches between training and testing distributions. The effect of the gate is stronger: injecting Toeplitz mixing without gating results in a large degradation (MAE 4.922 bpm), while the gated model achieves the best performance (MAE 1.582 bpm). This result suggests that gating is important for robustness under domain shift, because it provides a mechanism to limit the influence of global mixing when the learned temporal filter does not match the target domain characteristics.

TABLE V
CORE ABLATION ON SCAMPS→UBFC-rPPG

| Variant | MAE↓ | RMSE↓ | MAPE↓ | Pearson↑ | SNR↑ |
|---|---|---|---|---|---|
| Local-only | 2.285 | 3.871 | 3.071 | 0.992 | -4.738 |
| No gate | 4.922 | 8.963 | 5.875 | 0.974 | -5.533 |
| ToTMNet | 1.582 | 2.862 | 1.660 | 0.994 | -5.385 |

Two observations follow from these ablations. First, global Toeplitz mixing should be treated as a powerful operator that can improve rPPG extraction when used selectively, but it can also amplify non-physiological temporal structure if it is always active. Second, the gate provides a simple content-dependent control mechanism that makes global temporal mixing more reliable across conditions, which is particularly valuable for cross-dataset transfer where illumination statistics, motion patterns, and video rendering differ. In future work, it is worth studying whether more explicit reliability indicators (e.g., motion magnitude or ROI quality) can further improve the gating behavior.

## V. CONCLUSION

This paper presented ToTMNet, a lightweight rPPG architecture that replaces attention-based temporal modeling with FFT-accelerated Toeplitz temporal mixing. The Toeplitz operator provides global temporal receptive field with linear parameterization in clip length and efficient $O(T\log T)$ computation. ToTMNet combines local depthwise temporal convolution with gated global Toeplitz mixing to achieve strong heart-rate estimation accuracy while remaining compact. On UBFC-rPPG intra-dataset evaluation, ToTMNet achieves 1.055 bpm MAE, and it achieves 1.582 bpm MAE in a SCAMPS→UBFC-rPPG synthetic-to-real setting. Ablations confirm that the gating mechanism is critical to effectively utilize global Toeplitz mixing, especially under domain shift. The primary limitation of this preprint study is evaluation on only two datasets; extending ToTMNet to additional public datasets and reporting thorough efficiency benchmarks are important next steps for a journal submission.